\documentclass[conference]{IEEEtran}

\usepackage{amsmath,amssymb}
\usepackage{booktabs}
\usepackage{cite}
\usepackage{graphicx}
\usepackage{listings}
\usepackage{tikz}
\usepackage{url}
\usetikzlibrary{arrows.meta, positioning, shapes.geometric, backgrounds, fit, calc}
\newcommand{\ttbreak}{\discretionary{}{}{}}
\begin{document}

\title{Verification of PETSc with CIVL using
LLM-generated ACSL contracts and deterministic driver generation}

\author{%
  \IEEEauthorblockN{Hansol Suh\IEEEauthorrefmark{1},
    Jan H{\"u}ckelheim\IEEEauthorrefmark{1},
    Stephen Siegel\IEEEauthorrefmark{2}}
  \IEEEauthorblockA{\IEEEauthorrefmark{1}Argonne National Laboratory,
    Lemont, Illinois, USA\\
    \{hsuh,jhueckelheim\}@anl.gov\\
    ORCID: 0000-0001-8875-4074; 0000-0003-3479-6361}
  \IEEEauthorblockA{\IEEEauthorrefmark{2}University of Delaware,
    Newark, Delaware, USA\\
    siegel@udel.edu}
  \IEEEauthorblockA{Corresponding author: Hansol Suh}
}

\maketitle

\begin{abstract}
Parallel numerical libraries such as PETSc are widely used in science and engineering applications where wrong results can have costly consequences. Despite this, numerical libraries are rarely formally verified.
One of the challenges is the need for an expert to hand-write a specification and manually apply a verification tool, often requiring the development of a harness or driver, all of which can contain additional bugs that lead to false positives or false negatives during verification.
With recent advancements in large language models (LLMs), it is tempting to generate such drivers and reference models automatically,
but one-shot generation based on a simple prompt is brittle and leads to additional unverified code that needs to be audited.
In this paper, we present an approach to use LLMs in a limited setting to generate a small, human-certifiable ACSL contract from the function's documentation, combined with a deterministic toolchain that supports a restricted ACSL profile and generates a
driver that uses the CIVL verifier to check the implementation against the contract and, when available,
an existing reference model.
We demonstrate the pipeline end-to-end on three PETSc functions:
\texttt{MatAXPY} (reference model already exists),
\texttt{MatAYPX} (no reference model, so the certified contract is the sole
oracle), and the non-compressing mode of \texttt{MatFilter} (no reference
model, with more complex, conditional behavior).
With this pipeline, we were able to discover a bug in PETSc's
\texttt{MatAYPX} function that was previously undiscovered and had been present in the code since 1997.
\end{abstract}


\section{Introduction}

The Portable, Extensible Toolkit for Scientific Computation
(PETSc)~\cite{petsc-user-ref} is a widely used numerical
library in computational science, supplying parallel linear and
nonlinear solvers, ODE integrators, and optimization routines on which application
codes in domains as varied as aerodynamics, combustion, cardiology, and
subsurface flow are built. Because many high-consequence applications
depend on PETSc, it is tested extensively: its
continuous-integration pipelines exercise tens of thousands of tests on every
code change. Yet even high-coverage test suites can miss subtle defects, and a bug in
a core module, such as the vector (\texttt{Vec}) or matrix (\texttt{Mat}),
may silently compromise the correctness of many other functions that use them.
Formal verification can help detect bugs that occur only in corner cases and provide additional confidence in the computed results.
With the CIVL model checker~\cite{siegel2015civl}, one
writes, for a given function, a \emph{driver} that constructs symbolic
inputs, runs a CIVL-compatible port of the implementation, and compares the
result against a reference model; CIVL then checks that the two agree over all
inputs within bounds. 
Here, a CIVL-compatible port preserves the implementation paths being verified while
removing unsupported features, such as sparse storage.

In the past, CIVL has been used to verify parts of PETSc's vector module~\cite{dhavala2025verifying},
and it found two previously undiscovered bugs.
That result required hand-written drivers, reference models, and supporting
artifacts for each function, which does not scale to a code base the size of PETSc.

To address this scaling problem, large language models (LLMs) are an
obvious temptation:
PETSc has extensive documentation, which makes one-shot documentation-to-driver
generation via LLM plausible.
However, such an LLM-generated driver is itself unverified code.
Auditing the symbolic inputs, MPI setup, and oracle logic still requires the
same domain knowledge needed to write a driver, and an unaudited
driver that passes verification is worse than no driver at all, as it results in misplaced confidence in the underlying code.

To fully exploit the scalability of the LLM while minimizing the
LLM-written code that a human must manually inspect, we separate the stochastic output
of the LLM from the parts that can be produced deterministically.
Instead of asking the LLM to emit a full driver, we ask it only for a short
formal contract, written in a restricted ACSL profile, derived from
the function's documentation.
A human reviewer with PETSc expertise then validates the contract using the
documentation, API semantics, and implementation where necessary, recording
the rationale for any decisions not settled by the documentation. Only a
certified contract proceeds. To turn the certified contract into an
executable driver without reintroducing unverified code, we present
\texttt{acsl2civl}, a deterministic tool that translates the certified
contract into a CIVL verification driver.
This approach is less automated than generating the driver directly from the specification via an
LLM, but it concentrates human review on a short contract instead of requiring
the reviewer to audit an entire generated driver.
This paper contributes:
\begin{enumerate}
  \item the three-stage pipeline (LLM $\to$ manual review $\to$ deterministic
  compiler) to turn natural-language documentation into a machine-verifiable driver (\S\ref{sec:approach});
  \item a restricted PETSc-ACSL profile over a dense logical image of PETSc
          objects (\S\ref{subsec:repr}), and a manual review step for evolving
          the profile (\S\ref{subsec:profile-evolution});
  \item \texttt{acsl2civl}, a deterministic compiler that generates
  a CIVL driver from the human-certified contract. The driver checks the
  implementation against the contract's ACSL formulas and, when available,
  also checks agreement with a separately authored reference model (\S\ref{subsec:stage3});
  \item an evaluation on three operations: (i)~\texttt{MatAXPY}, verified against a specification and reference model (\S\ref{subsec:mataxpy}); (ii)~\texttt{MatAYPX}, verified only against a LLM-generated and human-checked contract and exposing the aforementioned PETSc bug
  (\S\ref{subsec:mataypx}); and (iii)~the non-compressing mode of
  \texttt{MatFilter}, which demonstrates a more complicated postcondition using conditionals (\S\ref{subsec:matfilter}).
\end{enumerate}

\subsection{Background on the CIVL model checker}
CIVL~\cite{siegel2015civl} is a symbolic execution and model checking
framework for CIVL-C, a superset of C with first-class concurrency and
verification primitives, with faithful semantics for a subset of MPI
\cite{luo-etal:2017:mpi}. Given a program with symbolic inputs,
CIVL explores its state space in order to
verify properties over all interleavings, within user-specified bounds on
inputs and the number of processes.  It discharges assertions with automated
theorem provers, reporting either a counterexample trace or that the chosen
fault classes are absent on all executions. The line of work behind CIVL includes
verification of collective operations~\cite{siegel2011collective} and
contract-based verification of MPI programs~\cite{luo-siegel:2024:contracts}.

To verify a library function $f$ with CIVL, one writes a \emph{verification driver}.
The driver constructs appropriate symbolic inputs for $f$, and
then calls both the actual implementation of $f$ and a trusted
reference model for $f$, when available. The driver
then compares the two results symbolically, reporting an error if it
fails to prove they are equivalent.
The driver and the reference model have, until now, been written by hand for
every function.

\subsection{PETSc}

PETSc~\cite{petsc-user-ref} is a widely used open-source library for
large-scale scientific computing on parallel, distributed-memory systems.
Its solvers and time integrators are built on core abstractions such as
\texttt{Vec} and \texttt{Mat}, whose implementations hide distributed storage,
assembly state, and type-specific kernels behind opaque handles and dynamic
dispatch. These features make PETSc a valuable verification target, but they
also make direct verification difficult: a driver must construct legal
symbolic objects, respect collective semantics, and observe the logical
mathematical object rather than incidental storage layout.

This work builds on an existing CIVL--PETSc verification
infrastructure~\cite{dhavala2025verifying}. That infrastructure provides an
abstract CIVL model of PETSc \texttt{Vec}/\texttt{Mat} objects and, for many
operations, three hand-written per-function artifacts: a reference model
\texttt{<Func>\_spec}, a CIVL-compilable port \texttt{<Func>.c} of the PETSc
interface function, and a driver that executes the port and checks its result
against the reference model. Our pipeline keeps the CIVL--PETSc object model
and PETSc porting discipline, but replaces the hand-written driver with one
generated deterministically from a human-certified contract. When no
reference model exists, as for \texttt{MatAYPX}, the certified contract becomes
the sole oracle checked by CIVL (\S\ref{subsec:stage3}).

\section{Approach}
\label{sec:approach}

\begin{figure}[t]
  \centering
\begin{tikzpicture}[
  font=\footnotesize\sffamily,
  >={Stealth[round]},
  node distance=4mm,
  box/.style={
    rectangle,
    rounded corners=2pt,
    draw=black!70,
    thick,
    align=center,
    text width=55mm,
    inner sep=4pt,
    fill=white
  },
  input/.style={box, fill=black!5},
  stage1/.style={box, fill=blue!7},
  stage2/.style={box, fill=orange!12},
  stage3/.style={box, fill=green!9},
  verify/.style={box, fill=black!5},
  decision/.style={
    diamond,
    aspect=2.5,
    draw=black!70,
    thick,
    align=center,
    inner sep=1pt,
    text width=15mm,
    fill=white
  },
  result/.style={input, rounded corners=8pt},
  flow/.style={->, thick, black!75},
  note/.style={font=\scriptsize\sffamily, text=black!65}
]

\node[input] (doc) {\textbf{PETSc documentation and prototype}};

\node[stage1, below=of doc] (llm) {
  \textbf{Stage 1: LLM (untrusted)}\\
  generate a candidate PETSc--ACSL contract\\
  using the supported ACSL and custom PETSc predicates
};

\node[stage2, below=of llm] (review) {
  \textbf{Stage 2: PETSc-aware human review}\\
  check documented behavior; resolve and record\\
  documentation gaps using relevant evidence
};

\node[decision, below=5mm of review] (certified) {certified?};

\node[stage3, below=6mm of certified] (compiler) {
  \textbf{Stage 3: \texttt{acsl2civl} (deterministic)}\\
  parse and check supported constructs; derive cases;\\
  generate CIVL driver(s)
};

\node[verify, below=of compiler] (civl) {
  \textbf{\texttt{civl verify}}\\
  check the CIVL-compatible PETSc port against\\
  contract assertions and, when available, a reference model
};

\node[result, below=of civl] (result) {
  \textbf{verified within bounds}\qquad or\qquad \textbf{counterexample}
};

\draw[flow] (doc) -- (llm);
\draw[flow] (llm) -- node[right, note]{candidate contract} (review);
\draw[flow] (review) -- (certified);
\draw[flow] (certified) -- node[right, note]{yes: certified contract} (compiler);
\draw[flow] (compiler) -- node[right, note]{generated driver(s)} (civl);
\draw[flow] (civl) -- (result);

\coordinate (feedback) at ($(review.west)+(-7mm,0)$);
\draw[flow] (certified.west) -- node[above, note]{no} (certified.west -| feedback)
  |- node[left, note, pos=0.25]{} (llm.west);

\end{tikzpicture}
  \caption{The three-stage pipeline: LLM-generated contract, human
  certification, and deterministic CIVL driver generation.}
  \label{fig:pipeline}
\end{figure}
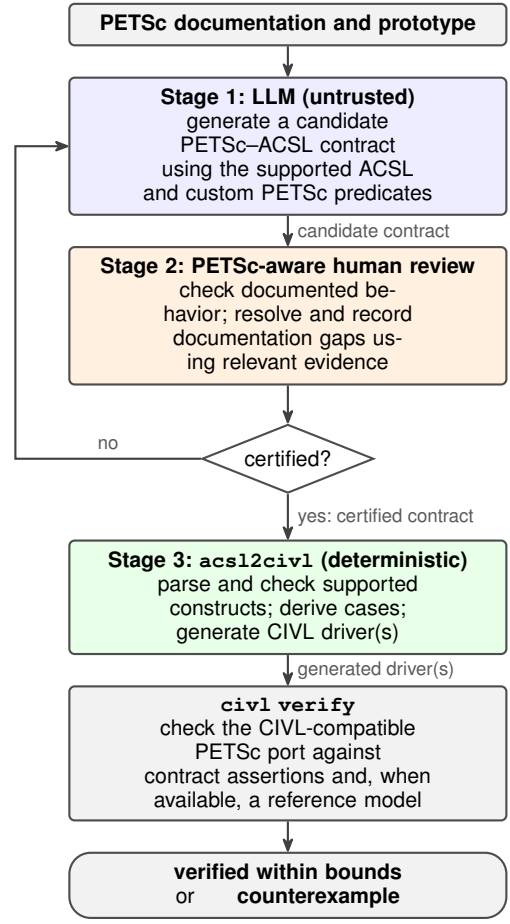

One-shot LLM driver generation bundles three distinct tasks:
(i) identifying the behavior documented by a PETSc manual page,
(ii) encoding that behavior in a symbolic contract,
and (iii) engineering the symbolic-MPI harness needed to check it.
The first two require language and modeling judgment, where using an LLM can be useful.
The third is mechanical and should be deterministic:
when generated by an LLM, it is also the hardest to audit, because a driver mixes object
construction, case splitting, frame checks, reference-model calls, and CIVL-specific plumbing.
Our design principle is therefore to \emph{separate the stochastic step from
the deterministic ones}, joined by manual inspection of a small,
reviewable artifact (Figure~\ref{fig:pipeline}):

\begin{enumerate}
  \item \textbf{Stage 1 (stochastic, LLM):} the LLM turns PETSc
  documentation into a small candidate contract in the fixed PETSc-ACSL
  profile (\S\ref{subsec:repr}); this contract is the only
  artifact the LLM generates.
  \item \textbf{Stage 2 (human certification):} a PETSc-aware expert
  checks the candidate against the documentation and, if necessary, resolves gaps in the documentation by reviewing the API and implementation, or asking the PETSc developers for clarification; failure returns the
  contract to Stage~1, and certification sends it to Stage~3.
  \item \textbf{Stage 3 (deterministic, no LLM):} \texttt{acsl2civl} translates
  the certified contract into CIVL driver(s) or rejects it if it contains unsupported constructs.
\end{enumerate}

CIVL then checks each generated driver against the
CIVL-compilable port of the PETSc implementation for the function being verified. This division
limits LLM output to the reviewed contract; every step afterwards is performed by deterministic, reviewable tools.

\subsection{Symbolic representation: a PETSc contract language}
\label{subsec:repr}

We express contracts using a supported fragment of
ACSL~\cite{baudin_acsl,kirchner2015framac}, a behavioral specification
language for C that provides clauses such as \lstinline|requires|,
\lstinline|assigns|, and \lstinline|ensures|.
We augment this fragment with the custom PETSc predicates listed in
Table~\ref{tab:petsc-symbols}.
\texttt{acsl2civl} interprets these extensions and translates the contracts
into CIVL-C driver code.
ACSL was chosen for
three reasons:
(i) ACSL allows the formulation of postconditions in a format understandable by the subsequent verification pipeline,
(ii) ACSL can be naturally added to C, which is the language PETSc is written in and that CIVL understands, so contracts can sit in documentation
comments beside the PETSc prototypes they describe and remain readable to both
the human reviewer and the remaining toolchain,
(iii) ACSL leaves a path to
Frama-C-based mechanical checks as future work (\S\ref{sec:related-work}).
While proof assistants such as Lean~\cite{moura2021lean} or Rocq~\cite{RocqProver}
may also be useful, the authors believe it is preferable to keep the contract easily accessible to PETSc developers, who are accustomed to C.

For verification, we model each distributed PETSc vector or matrix as a dense,
row-major global image, regardless of its actual storage format, written
\lstinline|M->data[k]| for
\lstinline|0 <= k < M->rmap->N * M->cmap->N|. Distributed storage or MPI communication are abstracted away.

The supported ACSL fragment is deliberately small: \lstinline|requires| /
\lstinline|assigns| / \lstinline|ensures| clauses (plus named
\lstinline|behavior|s as an escape hatch, \S\ref{subsec:profile-evolution});
terms over \lstinline|\old|, \lstinline|\forall|, scalar arithmetic,
\lstinline|\abs|, ordered comparisons, and conditional
expressions; \texttt{PetscScalar}, \texttt{PetscReal}, and pinned
\texttt{PetscBool} parameters; and object (in)equality
(\lstinline|==| / \lstinline|!=|). Reductions
(\lstinline|\sum|, norms, dot products) are not yet supported.

\begin{table}[t]
  \caption{Custom PETSc predicates for \texttt{acsl2civl}.
  These unescaped names extend the supported ACSL fragment.}
  \label{tab:petsc-symbols}
  \small
  \begin{tabular}{@{}p{0.32\linewidth}p{0.60\linewidth}@{}}
    \toprule
    \textbf{Symbol} & \textbf{Stipulated meaning in the PETSc model} \\
    \midrule
    \lstinline|valid_mat(M)| &
    $M$ is a matrix handle whose modeled dimensions, communicator,
    assembly state, and dense global image are consistent. \\
    \addlinespace
    \lstinline|valid_vec(v)| &
    $v$ is a vector handle whose modeled size, communicator,
    assembly state, and dense global image are consistent. \\
    \addlinespace
    \lstinline|separated_mat(A,B)| &
    $A$ and $B$ are distinct matrix objects with disjoint modeled storage. \\
    \addlinespace
    \lstinline|separated_vec(x,y)| &
    $x$ and $y$ are distinct vector objects with disjoint modeled storage. \\
    \addlinespace
    \lstinline|same_comm(x,y)| &
    The modeled PETSc objects use the same MPI communicator. \\
    \addlinespace
    \lstinline|assembled(x)| &
    The modeled object has no pending assembly updates. \\
    \addlinespace
    \texttt{logically\_\allowbreak collective(x,a)} &
    Corresponds to PETSc's \texttt{Petsc\ttbreak Valid\ttbreak Logical\ttbreak Collective\{Real,\ttbreak Scalar\}}.
    All ranks in $x$'s communicator provide the same value for $a$. \\
    \addlinespace
    \lstinline|scalar_conj(z)| &
    The complex conjugate of $z$; the identity for real \texttt{PetscScalar}. \\
    \bottomrule
  \end{tabular}
\end{table}

\subsection{Stage 1: from documentation to a candidate contract}
\label{subsec:stage1}

Stage 1 uses the LLM to turn a PETSc manual page into a candidate contract in
the profile of \S\ref{subsec:repr}. This contract is the only artifact the LLM
authors, and the LLM does not see the PETSc source code repository. This is done to avoid the generation of a contract that is influenced by (possibly erroneous) implementation choices. The LLM only views the
function's documentation, the PETSc function prototype, and the profile
definition of \S\ref{subsec:repr}. We acknowledge that, due to the popularity of PETSc, the LLM may nevertheless contain knowledge of PETSc source code, but we postulate that excluding the source code from the context and prompt will nevertheless increase the likelihood of obtaining specifications that are grounded in the natural-language documentation as opposed to the implementation.

The documentation supplies the behavioral content the contract is meant to
formalize: the operation's intended effect and the documented parameter semantics.
When the documentation does not specify behavior for a case, such as parameter aliasing,
the contract must make that gap and any resulting assumption explicit.
Stage 1 may be repeated multiple times based on feedback from Stages 2 and 3.

The Stage~1 prompt is a self-contained Markdown document
(\S\ref{sec:artifact}).
It contains the current profile definition, a generic output skeleton, and the
target function's documentation and prototype.
The profile evolves between runs as Stage~3 gains new capabilities.
Each prompt is piped to Claude Code using Claude Opus 4.8, with session
persistence, external setting sources, and tools disabled.\footnote{The exact
command was
  \texttt{\detokenize{cat $PROMPT_DIR/}}\allowbreak
  \texttt{\detokenize{Mat<Func>-prompt.md}}
\allowbreak\texttt{\detokenize{ | claude -p --model claude-opus-4-8}}
\allowbreak\texttt{\detokenize{ --no-session-persistence --setting-sources ""}}
  \allowbreak\texttt{\detokenize{ --allowed-tools}}\allowbreak
  \texttt{\detokenize{ ""}}.}
The CLI therefore has no repository access.
For each run, the complete prompt and resulting contract are archived with the
artifact; any corrections made during Stage~2 human review are reflected in the
certified contract.

\subsection{Stage 2: human-certification}
\label{subsec:gate}

A PETSc-aware expert reviews the candidate
contract against the documentation and relevant API and implementation. Behavioral formulas must match explicit documentation; where the
documentation is unclear, the reviewer must make and record a justified policy
decision rather than hiding it as an unstated assumption. Relevant
preconditions outside the profile are likewise recorded. For example,
\texttt{MatAXPY(Y, a, X, str)}'s manpage does not mention aliasing. A human reviewer can inspect the documentation of similar functions (such as \texttt{MatAYPX}), determine that aliasing is typically allowed and handled correctly, and record the decision to admit
\texttt{X == Y} as an allowable input.
The documentation also marks the operation logically collective, so the
certified contract records that obligation as
\lstinline|requires logically_collective(Y, a)|.
Rejected contracts are revised or regenerated before another review pass; accepted contracts are stamped
\texttt{CERTIFIED} with a revision number, date, and certifier, then released
to Stage~3.

Certification is complemented by the mechanical profile checks performed by
\texttt{acsl2civl} before driver generation (\S\ref{subsec:stage3}) to enforce syntactic well-formedness.

\subsection{Stage 3: deterministic translation to a CIVL driver}
\label{subsec:stage3}
Stage~3 is implemented by \texttt{acsl2civl} (written in Python using the Lark parsing toolkit). The tool
extracts the function signature and ACSL block, parses the contract into a
small specification IR, derives the admissible cases
(\S\ref{subsec:profile-evolution}), and emits one CIVL driver per case. It can
also emit Makefiles that are then used to run the
experiments and check expected results. There is no LLM in this stage.
Contracts that fail to run for any reason are rejected and passed back to the previous stages.

The generated drivers run a CIVL-compatible port \texttt{<Func>.c} of the
PETSc implementation under test.
The port preserves the implementation paths being verified while removing
currently unsupported features, such as sparse matrix data type.
Earlier CIVL--PETSc work manually implemented reference models for selected
operations.
Of our three examples, only \texttt{MatAXPY} has such a model.
For \texttt{MatAXPY}, the driver compares the implementation result with the
reference model result, and separately checks the contract-derived assertions.
No reference models exist for \texttt{MatAYPX} or \texttt{MatFilter}, so their
contract-derived assertions provide the correctness criteria.
Table~\ref{tab:lowering} summarizes how supported contract constructs are
translated into these checks.

\begin{table}[t]
  \caption{Translation from supported contract constructs to CIVL-C driver code.}
  \label{tab:lowering}
  \small
  \begin{tabular}{@{}p{0.32\linewidth}p{0.60\linewidth}@{}}
    \toprule
    \textbf{Contract construct} & \textbf{Generated driver behavior} \\
    \midrule
    Supported \lstinline|requires| equality &
    constrains symbolic dimensions or parameter values during driver
    construction. \\
    \addlinespace
    Custom PETSc predicate &
    emits corresponding CIVL assumptions, construction constraints, or
    generated cases. \\
    \addlinespace
    \lstinline|assigns| over modeled data &
    records pre-call values outside the assignable set and checks that those
    values remain unchanged after the call. \\
    \addlinespace
    Supported \lstinline|ensures| &
    translated into elementwise CIVL assertions over the gathered global
    image; \lstinline|\old| terms refer to recorded pre-call values. \\
    \bottomrule
  \end{tabular}
\end{table}

\subsection{Trust boundaries}
\label{subsec:trust}
For a generated verification driver, the key question is which artifacts are
trusted and which are checked.
The trusted pieces are
(i) CIVL itself,
(ii) the hand-written CIVL--PETSc module, 
(iii) CIVL-compatible ports \texttt{<Func>.c}, whose fidelity to PETSc source is trusted, and
(iv) \texttt{acsl2civl}.
The LLM is untrusted. The human-certified contract is trusted to state the
intended behavior. With \texttt{--contract-assert}, CIVL checks both the
contract and the reference model. With \texttt{--no-spec}, as for
\texttt{MatAYPX}, the contract postconditions and frame conditions are the only
correctness checks.

\section{Evaluation}
\label{sec:evaluation}

We study three successive Stage~1 runs.
\texttt{MatAXPY} establishes the baseline contract form and motivates derived
aliasing cases; \texttt{MatAYPX} applies that extension in contract-only mode;
and the non-compressing mode of \texttt{MatFilter} exercises a conditional
postcondition.
Unless stated otherwise, the reported \texttt{small} experiments exhaustively
explore all executions and symbolic inputs with MPI process counts 1 and 2 and global matrix dimensions $M,N$ from
1 to 3 within the CIVL--PETSc model and the modes admitted by each certified
contract.

Table~\ref{tab:civl-performance} reports fresh measurements.\footnote{Measured
on a laptop with an Intel Core Ultra 7 258V processor and 32\,GiB of memory,
using OpenJDK 21.0.11, CIVL 2.0 (2026-07-16), Z3 4.8.12, and CVC4 1.8.}
For \texttt{MatAXPY} and \texttt{MatAYPX}, \texttt{D} and \texttt{U} denote
\texttt{DIFFERENT\_NONZERO\_PATTERN} and \texttt{UNKNOWN\_NONZERO\_PATTERN},
respectively.
The aliased \texttt{MatAYPX} rows terminate at the provable
contract violation.

\begin{table}[tb]
  \caption{CIVL verification cost for the \texttt{small} experiments.}
  \label{tab:civl-performance}
  \centering
  \scriptsize
  \setlength{\tabcolsep}{2.5pt}
  \begin{tabular*}{\columnwidth}{@{\extracolsep{\fill}}lllrrrr@{}}
    \toprule
    Function & Case & \texttt{str} & Time (s) & Mem. (GiB) & States & Trans. \\
    \midrule
    \texttt{MatAXPY} & aliased & D & 27.01 & 2.27 & 79,599 & 162,530 \\
    \texttt{MatAXPY} & aliased & U & 29.09 & 2.36 & 79,599 & 162,530 \\
    \texttt{MatAXPY} & distinct & D & 36.73 & 2.27 & 102,082 & 206,215 \\
    \texttt{MatAXPY} & distinct & U & 34.36 & 2.27 & 102,082 & 206,215 \\
    \texttt{MatAYPX} & aliased$^\dagger$ & D & 6.75 & 1.30 & 1,615 & 3,460 \\
    \texttt{MatAYPX} & aliased$^\dagger$ & U & 7.06 & 1.71 & 1,615 & 3,460 \\
    \texttt{MatAYPX} & distinct & D & 29.16 & 2.27 & 71,429 & 149,142 \\
    \texttt{MatAYPX} & distinct & U & 29.72 & 2.27 & 71,429 & 149,142 \\
    \texttt{MatFilter} & --- & --- & 18.29 & 1.30 & 20,502 & 40,062 \\
    \bottomrule
  \end{tabular*}
  \\[2pt]
  \parbox{0.96\columnwidth}{\footnotesize $^\dagger$assertion violation for the aliasing bug.}
\end{table}

Table~\ref{tab:review-size} compares three review strategies by counting
lines of code, excluding blank and comment-only lines.
The columns count the original PETSc body, its CIVL-compatible port, all
generated case drivers, and the ACSL clauses in the certified contract.

\begin{table}[tb]
        \caption{Artifact size by review strategy.}
  \label{tab:review-size}
  \centering
  \small
  \begin{tabular*}{\columnwidth}{@{\extracolsep{\fill}}lcccc@{}}
    \toprule
    Function & PETSc body & Port & Drivers & Contract \\
    \midrule
    \texttt{MatAXPY} & 39 & 23 & 176 & 14 \\
    \texttt{MatAYPX} & 4 & 4 & 152 & 13 \\
    \texttt{MatFilter} & 72 & 24 & 38 & 8 \\
    \bottomrule
  \end{tabular*}
\end{table}

These counts do not measure review time.
They nevertheless expose the immediate review surface: 8--14 contract lines
instead of 38--176 generated-driver lines per function.

\subsection{Worked example: \texttt{MatAXPY}}
\label{subsec:mataxpy}

Consider the PETSc documentation for \texttt{MatAXPY}, excerpted in
Listing~\ref{lst:mataxpy-doc}. It states the value update and lists the
\texttt{str} nonzero-pattern hints, but it does not specify whether the input
matrices may alias.

\begin{lstlisting}[float=tb,floatplacement=tb,caption={PETSc documentation excerpt for \texttt{MatAXPY}.},label={lst:mataxpy-doc},captionpos=b]
MatAXPY - Computes Y = a*X + Y.

Logically Collective

Input Parameters:
+ a   - the scalar multiplier
. X   - the first matrix
. Y   - the second matrix
- str - either SAME_NONZERO_PATTERN,
        DIFFERENT_NONZERO_PATTERN,
        UNKNOWN_NONZERO_PATTERN, or
        SUBSET_NONZERO_PATTERN
        (nonzeros of X is a subset of Y's)
\end{lstlisting}

The \texttt{str} argument is a nonzero-pattern hint used by PETSc for internal
optimization. The contract admits \texttt{DIFFERENT\_\ttbreak NONZERO\_\ttbreak PATTERN} and
\texttt{UNKNOWN\_\ttbreak NONZERO\_\ttbreak PATTERN}, and Stage~3 tests both with distinct and
aliased handles.
We exclude \texttt{SAME\_\ttbreak NONZERO\_\ttbreak PATTERN} and
\texttt{SUBSET\_\ttbreak NONZERO\_\ttbreak PATTERN} because the current dense model cannot represent or check their structural preconditions.
\texttt{DIFFERENT\_\ttbreak NONZERO\_\ttbreak PATTERN} is not meaningful
when $X = Y$, but PETSc's same-handle path ignores the hint entirely. Stage~3 therefore retains this case rather than imposing an
undocumented restriction. The dense model checks the resulting matrix values,
not the sparse-pattern claim itself.

A clean-context Stage-1 run for
\texttt{MatAXPY} produces the contract of Listing~\ref{lst:mataxpy}. The LLM
transcribes the value relation $Y = aX + Y$ over the dense image and, finding
the documentation silent on whether $X = Y$ (aliasing) is allowed, takes a conservative
position: it \emph{excludes} aliasing with a
\lstinline|requires separated_mat(X, Y)| clause and records that choice for the
reviewer. Its certification notes go further: they observe that the value relation
still holds under aliasing (\lstinline|\old| captures the pre-state, so
$X = Y$ gives $Y = (1+a)Y$) and that the \lstinline|separated_mat| restriction is
therefore model-generated, not documented.

The formula holding under aliasing, however, does not settle the question:
$X = Y$ is a different code path, not a special value of the same one, because the
\texttt{MatAXPY} implementation branches on the aliased case with
\lstinline|if (Y == X) MatScale(Y, 1+a)| (a comparison of pointers / memory addresses, not values). A driver over symbolic matrix values
never reaches that branch, because it allocates $X$ and $Y$ as two objects, so the
pointer test \lstinline|Y == X| is false regardless of the values CIVL explores.

Only a same-handle driver for $X$ and $Y$ exercises it. We therefore treat the
issue as a profile/tooling gap rather than asking the LLM to generate an
aliased case not specified by the documentation.
The fix is to encode two aliasing cases in the PETSc-ACSL profile: the same handle and
distinct handles with disjoint storage. We then update \texttt{acsl2civl} to derive
those cases from a contract.
With that derivation in place, the human reviewer certifies a
revised contract with the Stage-1-generated \lstinline|separated_mat| clause
removed. \texttt{acsl2civl} then generates one driver for each combination of
aliasing case and admitted \texttt{str} value.
Listing~\ref{lst:mataxpy} shows the raw Stage-1 proposal; the artifact contains
the corrected and certified Stage-2 contract.

\begin{lstlisting}[float=tb,floatplacement=tb,caption={Stage-1 output for \texttt{MatAXPY}},label={lst:mataxpy},captionpos=b]
/*@ requires valid_mat(Y);
    requires valid_mat(X);
    requires assembled(Y);
    requires assembled(X);
    requires same_comm(X, Y);
    requires (*@\colorbox{yellow!30}{\texttt{separated\_mat(X, Y)}}@*);
    requires Y->rmap->N == X->rmap->N;
    requires Y->cmap->N == X->cmap->N;
    requires logically_collective(Y, a);
    requires str == DIFFERENT_NONZERO_PATTERN
          || str == UNKNOWN_NONZERO_PATTERN;
    assigns Y->data[0 .. Y->rmap->N*Y->cmap->N - 1];
    ensures \forall integer i;
      0 <= i < Y->rmap->N*Y->cmap->N ==>
      Y->data[i] == a * \old(X->data[i])
                  + \old(Y->data[i]);
    ensures \result == 0; @*/
PetscErrorCode MatAXPY(Mat Y, PetscScalar a,
                       Mat X, MatStructure str);
\end{lstlisting}

Once the \lstinline|separated_mat| clause is removed, Stage~3 generates four
drivers: \texttt{distinct\_\ttbreak different}, \texttt{distinct\_\ttbreak unknown},
\texttt{aliased\_\ttbreak different}, and \texttt{aliased\_\ttbreak unknown},
where \texttt{distinct} means $X$, $Y$ are different objects while \texttt{aliased} says they
are the same. Here, \texttt{different} and \texttt{unknown} abbreviate the
corresponding \texttt{str} values in Listing~\ref{lst:mataxpy-doc}.
Listing~\ref{lst:mataxpy-driver} shows the core of the
\texttt{distinct\_\ttbreak different} driver: a reference-model oracle call, a call to
the ported PETSc implementation, whole-image comparison against the reference
result, an assigns-derived frame check on $X$, and certified postcondition
assertions enabled by \texttt{--contract-assert}.

\begin{lstlisting}[float=tb,floatplacement=tb,caption={Excerpt of the generated \texttt{distinct\_different} driver.},label={lst:mataxpy-driver},captionpos=b]
MatAXPY_spec(exp_Y, a, p_X, DIFFERENT_NONZERO_PATTERN);
PetscErrorCode err = MatAXPY(p_Y,a,p_X,DIFFERENT_NONZERO_PATTERN);
$assert(err == 0);
debug_print_matrix_result(ctx, p_Y, exp_Y);

verify_matrix_result(ctx, p_Y, exp_Y,
  "MatAXPY [distinct_different] postcondition on Y");

verify_matrix_result(ctx, p_X, exp_X,
  "MatAXPY [distinct_different] frame on X");

/* certified-contract assertions (--contract-assert mode): */
$mat img_Y = CIVL_PetscToCivlMat(p_Y);
$mat img_X = CIVL_PetscToCivlMat(p_X);
if (ctx.rank == 0) {
  STYPE X_pre[M * N];
  STYPE Y_pre[M * N];

  for (int k = 0; k < M * N; k++) {
    X_pre[k] = scalar_make(X_Real[k],IMAG_SCALAR_INPUT_IDX(X, k));
    Y_pre[k] = scalar_make(Y_Real[k],IMAG_SCALAR_INPUT_IDX(Y, k));
  }

  for (int k = 0; k < (M * N); k++)
    $assert(scalar_eq(img_Y.data[k],
      scalar_add(scalar_mul(a, X_pre[k]), Y_pre[k])),
      "MatAXPY [distinct_different] certified ensures #1
       on Y violated");

  for (int k = 0; k < M * N; k++)
    $assert(scalar_eq(img_X.data[k], X_pre[k]),
      "MatAXPY [distinct_different] certified frame
       on X violated");
}
\end{lstlisting}

The other functions studied in this paper ---
\texttt{MatAYPX} and \texttt{MatFilter} --- follow the same artifact pattern,
so we do not include their full contracts and generated drivers in the main text.
Their prompts, raw Stage~1 outputs, certified contracts, generated drivers, and
CIVL-compatible ports are included in the accompanying artifact
(\S\ref{sec:artifact}).

\subsection{Evolving the profile}
\label{subsec:profile-evolution}

The PETSc-ACSL profile is extensible through a human-in-the-loop process.
When the same contract structure appears repeatedly, we move it into the
profile and let Stage~3 generate it mechanically.
We illustrate this refinement loop with the \texttt{MatAXPY} aliasing split.

\begin{enumerate}
  \item \textbf{Inspect.} The clean-context Stage~1 run emits one contract
  (Listing~\ref{lst:mataxpy}).
  It excludes $X = Y$ with \lstinline|separated_mat(X, Y)|, but notes that the
  \lstinline|\old|-based postcondition remains coherent under $X = Y$, thereby
  flagging the ambiguity without inventing a case split.
  \item \textbf{Identify.} Retaining the clause would leave the
  imple\-mentation-supported $X = Y$ branch unverified.
  \item \textbf{Decide, not re-prompt.} Because the manpage is silent on
  aliasing, re-prompting would ask the LLM to invent undocumented semantics.
  Based on implementation evidence, human intervention is required to decide the correct action, in this case removal of the generated
  \lstinline|separated_mat| clause, admitting $X = Y$ without authoring a separate
  behavior.
  \item \textbf{Mechanize.} A developer extends \texttt{acsl2civl} with an
  aliased-or-distinct derivation rule.
  Stage~3 then generates a distinct, disjoint-storage driver and an aliased
  ($X = Y$) driver from the amended contract.
\end{enumerate}

Further applications may identify additional recurring patterns suitable for explicit
Stage~3 rules, making \texttt{acsl2civl} more capable and reusable.

\begin{figure*}[t]
  \centering
\begin{tikzpicture}[
  font=\scriptsize\sffamily,
  >={Stealth[round]},
  node distance=5mm,
  flowbox/.style={
    rectangle,
    rounded corners=2pt,
    draw=black!70,
    thick,
    align=left,
    text width=34mm,
    minimum height=20mm,
    inner sep=4pt,
    fill=white
  },
  need/.style={flowbox, fill=orange!12},
  language/.style={flowbox, fill=blue!7},
  compiler/.style={flowbox, fill=green!9},
  output/.style={flowbox, fill=black!5},
  flow/.style={->, thick, black!75},
  feedback/.style={->, thick, dashed, black!55},
  note/.style={font=\scriptsize\sffamily, text=black!60, align=center}
]

\node[need, text width=34mm] (need) {
  \textbf{1. Identify unsupported semantics}\\[-1pt]
  \texttt{MatFilter} requires a symbolic real tolerance and the conditional update\\
  $|A_{\mathrm{old}}[i]| \leq \texttt{tol}\ ?\ 0 : A_{\mathrm{old}}[i]$.
};

\node[language, text width=40mm, right=of need] (language) {
  \textbf{2. Extend profile}\\[-1pt]
  \textbf{Profile:} allow symbolic \texttt{PetscReal tol}.\\
  \textbf{Parser:} recognize \texttt{\textbackslash abs}, ordered comparisons, and conditional expressions.\\
};

\node[compiler, text width=39mm, right=of language] (lower) {
  \textbf{3. Update \texttt{acsl2civl}}\\[-1pt]
  represent new operations in a form that \texttt{acsl2civl} can understand.\\
  emit \texttt{tol} as a symbolic CIVL input.\\
  \texttt{\textbackslash abs(e)} $\mapsto$ \texttt{scalar\_abs(e)}.\\
  support comparison and conditional expression in the generated assertion.
};

\draw[flow] (need) -- (language);
\draw[flow] (language) -- (lower);

\end{tikzpicture}
  \caption{Extending the PETSc-ACSL profile and \texttt{acsl2civl} to support
  \texttt{MatFilter}.}
  \label{fig:matfilter-profile-extension}
\end{figure*}
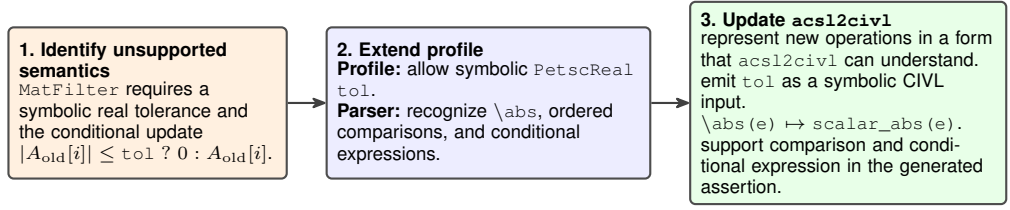

\subsection{Contract-only verification: \texttt{MatAYPX} and an aliasing bug}
\label{subsec:mataypx}

\texttt{MatAYPX} ($Y \gets aY + X$) has no reference model in the
CIVL--PETSc module, so its certified contract serves as the only oracle.
Because the documentation does not require $X$ and $Y$ to differ, the evolved
profile leaves aliasing unrestricted.
Stage~2 corrected the missing \lstinline|\old| in the postcondition. It also
expanded the tested \lstinline|str| values from
\texttt{UNKNOWN\_\ttbreak NONZERO\_\ttbreak PATTERN} to include
\texttt{DIFFERENT\_\ttbreak NONZERO\_\ttbreak PATTERN}, ensuring that every hint admitted by the
certified contract is exercised. The dense model checks only the numerical
result, not the associated sparse-pattern claim.
The manual review certified the amended contract, from which Stage~3 derived distinct
and aliased drivers.

The distinct drivers pass in contract-only mode, but the aliased drivers
produce a \textsc{provable} assertion violation.
Listing~\ref{lst:mataypx-impl} shows the PETSc implementation: it scales $Y$
and then adds $X$ to the result.

\begin{lstlisting}[float=tb,floatplacement=tb,caption={PETSc implementation of \texttt{MatAYPX}.},label={lst:mataypx-impl},captionpos=b]
PetscErrorCode MatAYPX(Mat Y, PetscScalar a, Mat X,
                       MatStructure str)
{
  PetscFunctionBegin;
  PetscCall(MatScale(Y, a));
  PetscCall(MatAXPY(Y, 1.0, X, str));
  PetscFunctionReturn(PETSC_SUCCESS);
}
\end{lstlisting}

When $X = Y$, the second call reads the already-scaled matrix and produces
$2aY_{\mathrm{old}}$ instead of the documented $(1+a)Y_{\mathrm{old}}$.
In contrast, \texttt{MatAXPY} explicitly handles aliased inputs and
satisfies its documented equation.
Because \texttt{MatAYPX} documents no aliasing restriction, its inconsistent
aliased result is a bug, not a reason to weaken the contract with
\lstinline|requires Y != X|.

\subsection{A conditional operation: \texttt{MatFilter}}
\label{subsec:matfilter}

We verify the dense, and non-compressing mode of
\texttt{MatFilter(A,\ttbreak\,tol,\ttbreak\,compress,\ttbreak\,keep)}, which zeroes every entry of $A$
whose magnitude is at most \texttt{tol} and preserves every other entry.
The two Boolean parameters \texttt{compress} and \texttt{keep} indicate whether
to compress the matrix and keep the diagonal entries after filtering.
We do not yet support those modes, so we set both parameters to false.
To express and translate this behavior, we added support for symbolic
\texttt{PetscReal} parameters, \lstinline|\abs|, relational comparisons, and
conditional expressions.
Figure~\ref{fig:matfilter-profile-extension} shows the steps we took to extend
the profile and update \texttt{acsl2civl} to support \texttt{MatFilter}.

A clean-context Stage~1 run produced the elementwise postcondition
\lstinline|A->data[i] ==|\allowbreak
\lstinline|(\abs(\old(A->data[i])) <= tol ?|\allowbreak
\lstinline| 0 : \old(A->data[i]))|,
which Stage~3 lowers directly to an elementwise CIVL-C assertion.
The generated driver treats \texttt{tol} as a symbolic input.

\section{Scope and limitations}
\label{sec:limits}
\paragraph{Assembly and model fidelity}
\label{subsec:assembly}
Elementwise algebraic routines such as \texttt{MatAXPY} admit direct relations
between the pre- and post-state logical images.
Assembly is harder because calls such as \texttt{VecSetValues()} and
\texttt{MatSetValues()} may apply locally owned updates immediately while
placing off-process updates in \texttt{VecStash} or \texttt{MatStash} objects.
The globally assembled object is established only after the corresponding
\texttt{AssemblyBegin}/\allowbreak\texttt{AssemblyEnd} pair communicates and combines
those pending updates.

The CIVL--PETSc module implements the scalar \texttt{VecStash} path needed to
exercise \texttt{VecAssemblyBegin}/\allowbreak\texttt{VecAssemblyEnd}, including
off-process communication and insertion-mode reconciliation.
Extending the same support to matrices is ongoing work.
Dense matrices are comparatively direct, but sparse matrices couple the
stash protocol to row-wise sparse storage, preallocation, duplicate entries,
and insertion-mode rules, making a faithful \texttt{MatStash} model
substantially more difficult to specify.

Reimplementing PETSc's stash and assembly machinery in CIVL duplicates complex
state and communication logic, creating another place for bugs that could mask
or manufacture verification failures.
Such model code therefore requires its own focused drivers and cross-checks,
enlarging the codebase that must be carefully reviewed and trusted.
Until that validation is complete, matrix insertion and assembly remain outside
the verification claim.

\section{Related Work}
\label{sec:related-work}

Recent work has investigated whether LLMs can translate natural-language
intent into formal specifications.
Endres et al.~\cite{endres2024nl2postcond} evaluate generated postconditions
and show both the promise of the approach and the need for semantic validation.
Cao et al.~\cite{cao2025informal} decompose translation from natural-language
requirements into six formal-verification tasks and provide training and
evaluation data spanning five languages, including ACSL.
Clover~\cite{sun2024clover} places generation in a closed loop with a formal
verifier, using verification feedback to improve generated programs and proofs.
ATLAS~\cite{baksys2026atlas} similarly synthesizes Dafny contracts, programs,
and proofs from programming problems, reference implementations, and tests;
it uses test-derived checks to detect inconsistent or weak specifications.
Our pipeline shares the goal of using formal tooling to constrain stochastic
generation, but addresses existing PETSc implementations and assigns the LLM a
narrower role: it proposes a compact contract from library documentation that a
domain expert certifies before any driver is generated.
The LLM output is never executed as a verification harness; after human
certification, the trusted deterministic compiler generates the driver.

ACSL provides C-level contracts, and Frama-C parses and analyzes those contracts
~\cite{baudin_acsl,kirchner2015framac}.
Our PETSc-ACSL profile deliberately supports only the fragment for which
\texttt{acsl2civl} has explicit lowering rules.
This is not general contract inference or full ACSL verification.
It is a restricted interface between natural-language PETSc documentation and
bounded CIVL models, with unsupported constructs rejected rather than
approximated.

A Frama-C plug-in could replace part of the custom profile parser.
The plug-in could reuse Frama-C's ACSL front end to check syntax, typing,
binding, and profile membership, then export a valid contract
representation for \texttt{acsl2civl}.
This would replace part of the current custom parsing path with checks grounded
in the ACSL implementation itself and make profile violations easier to
diagnose before human certification.
It would not eliminate the human review: Frama-C can establish that a contract is
well formed, but not that it faithfully captures incomplete PETSc documentation
or that a reviewer made the right policy decision for an undocumented case.

\section{Conclusions}

We presented a disciplined pipeline for LLM-assisted verification of PETSc
operations: the LLM generates only a small behavioral contract in a restricted
ACSL profile, a PETSc-aware human reviewer checks documented behavior and records
justified resolutions of documentation gaps, and a deterministic compiler
lowers the certified contract to CIVL drivers.
The contract and an optional, separately
authored reference model contain the behavior specification.
The design confines direct review of
LLM output to a compact contract, and the tool refuses anything outside its
profile rather than guessing.

Our evaluation covers \texttt{MatAXPY}, \texttt{MatAYPX}, and the
non-compressing mode of \texttt{MatFilter}.
Applying the pipeline across these routines also drove two extensions to
\texttt{acsl2civl}: derived aliasing cases and, for \texttt{MatFilter}'s
non-compressing mode, conditional postconditions over real parameters.
The derived aliased case exposed a bug in
\texttt{MatAYPX} that had remained undetected since 1997, which produces $2aY$ rather than $(1+a)Y$ when its operands are the same handle.

Future work will extend matrix assembly support to \texttt{MatStash} and sparse
storage; broaden the profile to additional \texttt{Mat} operations and
\texttt{Vec} reductions; lower supported \lstinline|requires| clauses into
driver assumptions; and develop the Frama-C profile-checking plug-in described
in \S\ref{sec:related-work}.

\appendices

\section{Artifact and Reproducibility}
\label{sec:artifact}

The artifact records the three versioned Stage~1 prompts---one for each profile
iteration---and the resulting contracts, corrected as needed during the
Stage~2 human review and then certified.
Contracts were authored with Claude (Anthropic);
the human review stamps (revision, date, certifier) are in each contract's
header. The Stage-3 compiler (\texttt{tools/acsl2civl}, Python + Lark) and
the CIVL--PETSc module it presupposes are part of the same repository.

\section{Artifact Availability}
The artifact is submitted as supplementary material with the paper.
A public archival URL will be added to the camera-ready version.

\section*{Acknowledgment}
This material is based upon work supported by the U.S.\ National
Science Foundation under Award Number CCF-2446130, and by the U.S.\
Department of Energy, Office of Science, Advanced Scientific Computing
Research (ASCR) Program, under Award Number DE-SC0025953.
Portions of this paper were edited with assistance from GPT 5.6 Sol, and Claude Opus 4.8.

{\small This report was prepared as an account of work sponsored by an agency
of the United States Government. Neither the United States Government
nor any agency thereof, nor any of their employees, makes any
warranty, express or implied, or assumes any legal liability or 
responsibility for the accuracy, completeness, or usefulness of any
information, apparatus, product, or process disclosed, or represents
that its use would not infringe privately owned rights. Reference
herein to any specific commercial product, process, or service by
trade name, trademark, manufacturer, or otherwise does not necessarily
constitute or imply its endorsement, recommendation, or favoring by
the United States Government or any agency thereof. The views and
opinions of authors expressed herein do not necessarily state or
reflect those of the United States Government or any agency thereof.}

\bibliographystyle{IEEEtran}
\bibliography{civl-petsc}

\begin{thebibliography}{10}
\providecommand{\url}[1]{#1}
\csname url@samestyle\endcsname
\providecommand{\newblock}{\relax}
\providecommand{\bibinfo}[2]{#2}
\providecommand{\BIBentrySTDinterwordspacing}{\spaceskip=0pt\relax}
\providecommand{\BIBentryALTinterwordstretchfactor}{4}
\providecommand{\BIBentryALTinterwordspacing}{\spaceskip=\fontdimen2\font plus
\BIBentryALTinterwordstretchfactor\fontdimen3\font minus
  \fontdimen4\font\relax}
\providecommand{\BIBforeignlanguage}[2]{{%
\expandafter\ifx\csname l@#1\endcsname\relax
\typeout{** WARNING: IEEEtran.bst: No hyphenation pattern has been}%
\typeout{** loaded for the language `#1'. Using the pattern for}%
\typeout{** the default language instead.}%
\else
\language=\csname l@#1\endcsname
\fi
#2}}
\providecommand{\BIBdecl}{\relax}
\BIBdecl

\bibitem{petsc-user-ref}
\BIBentryALTinterwordspacing
S.~Balay, S.~Abhyankar, M.~F. Adams, S.~Benson, J.~Brown, P.~Brune,
  K.~Buschelman, E.~M. Constantinescu, L.~Dalcin, A.~Dener \emph{et~al.},
  ``Petsc/tao users manual revision 3.25,'' Argonne National Laboratory (ANL),
  Argonne, IL (United States), Tech. Rep., 03 2026. [Online]. Available:
  \url{https://www.osti.gov/biblio/3025790}
\BIBentrySTDinterwordspacing

\bibitem{siegel2015civl}
S.~F. Siegel, M.~Zheng, Z.~Luo, T.~K. Zirkel, A.~V. Marianiello, J.~G.
  Edenhofner, M.~B. Dwyer, and M.~S. Rogers, ``Civl: the concurrency
  intermediate verification language,'' in \emph{Proceedings of the
  International Conference for High Performance Computing, Networking, Storage
  and Analysis}, 2015, pp. 1--12.

\bibitem{dhavala2025verifying}
V.~Dhavala, J.~H{\"u}ckelheim, P.~D. Hovland, and S.~F. Siegel, ``Verifying
  petsc vector components using civl,'' in \emph{International Conference on
  Computer Aided Verification}.\hskip 1em plus 0.5em minus 0.4em\relax
  Springer, 2025, pp. 148--161.

\bibitem{luo-etal:2017:mpi}
Z.~Luo, M.~Zheng, and S.~F. Siegel, ``Verification of mpi programs using
  civl,'' in \emph{Proceedings of the 24th European MPI Users' Group Meeting},
  2017, pp. 1--11.

\bibitem{siegel2011collective}
S.~F. Siegel and T.~K. Zirkel, ``Collective assertions,'' in
  \emph{International Workshop on Verification, Model Checking, and Abstract
  Interpretation}.\hskip 1em plus 0.5em minus 0.4em\relax Springer, 2011, pp.
  387--402.

\bibitem{luo-siegel:2024:contracts}
Z.~Luo and S.~F. Siegel, ``Collective contracts for message-passing parallel
  programs,'' in \emph{Computer Aided Verification (CAV 2024)}, ser. Lecture
  Notes in Computer Science, A.~Gurfinkel and V.~Ganesh, Eds., vol.
  14682.\hskip 1em plus 0.5em minus 0.4em\relax Cham: Springer, 2024, pp.
  44--68.

\bibitem{baudin_acsl}
P.~Baudin, J.-C. Filli{\^a}tre, C.~March{\'e}, B.~Monate, Y.~Moy, and
  V.~Prevosto, ``Acsl: Ansi/iso c specification,'' \emph{URL https://frama-c.
  com/html/acsl. html}, 2021.

\bibitem{kirchner2015framac}
F.~Kirchner, N.~Kosmatov, V.~Prevosto, J.~Signoles, and B.~Yakobowski,
  ``Frama-c: A software analysis perspective,'' \emph{Formal aspects of
  computing}, vol.~27, no.~3, pp. 573--609, 2015.

\bibitem{moura2021lean}
L.~d. Moura and S.~Ullrich, ``The lean 4 theorem prover and programming
  language,'' in \emph{International Conference on Automated Deduction}.\hskip
  1em plus 0.5em minus 0.4em\relax Springer, 2021, pp. 625--635.

\bibitem{RocqProver}
\BIBentryALTinterwordspacing
T.~R.~D. Team, ``The rocq prover,'' Sep. 2025. [Online]. Available:
  \url{https://doi.org/10.5281/zenodo.17473943}
\BIBentrySTDinterwordspacing

\bibitem{endres2024nl2postcond}
\BIBentryALTinterwordspacing
M.~Endres, S.~Fakhoury, S.~Chakraborty, and S.~K. Lahiri, ``Can large language
  models transform natural language intent into formal method postconditions?''
  \emph{Proc. ACM Softw. Eng.}, vol.~1, no. FSE, Jul. 2024. [Online].
  Available: \url{https://doi.org/10.1145/3660791}
\BIBentrySTDinterwordspacing

\bibitem{cao2025informal}
J.~Cao, Y.~Lu, M.~Li, H.~Ma, H.~Li, M.~He, C.~Wen, L.~Sun, H.~Zhang, S.~Qin
  \emph{et~al.}, ``From informal to formal--incorporating and evaluating llms
  on natural language requirements to verifiable formal proofs,'' in
  \emph{Proceedings of the 63rd Annual Meeting of the Association for
  Computational Linguistics (Volume 1: Long Papers)}, 2025, pp.
  26\,984--27\,003.

\bibitem{sun2024clover}
C.~Sun, Y.~Sheng, O.~Padon, and C.~Barrett, ``Clover: Closed-loop ver ifiable
  code generation,'' in \emph{International Symposium on AI
  Verification}.\hskip 1em plus 0.5em minus 0.4em\relax Springer, 2024, pp.
  134--155.

\bibitem{baksys2026atlas}
\BIBentryALTinterwordspacing
M.~Baksys, S.~Zetzsche, O.~Bouissou, R.~Delmas, S.~Kong, and S.~B. Holden,
  ``Atlas: Automated toolkit for large-scale verified code synthesis,'' 2026.
  [Online]. Available: \url{https://arxiv.org/abs/2512.10173}
\BIBentrySTDinterwordspacing

\end{thebibliography}

\end{document}